# Anomaly Detection in Automatic Generation Control Systems Based on Traffic Pattern Analysis and Deep Transfer Learning

Tohid Behdadnia *Student Member, IEEE,*, Geert Deconinck, *Senior Member IEEE*

*Abstract*—In modern highly interconnected power grids, automatic generation control (AGC) is crucial in maintaining the stability of the power grid. The dependence of the AGC system on the information and communications technology (ICT) system makes it vulnerable to various types of cyber-attacks. Thus, information flow (IF) analysis and anomaly detection became paramount for preventing cyber attackers from driving the cyber-physical power system (CPPS) to instability. In this paper, the ICT network traffic rules in CPPSs are explored and the frequency domain features of the ICT network traffic are extracted, basically for developing a robust learning algorithm that can learn the normal traffic pattern based on the ResNeSt convolutional neural network (CNN). Furthermore, to overcome the problem of insufficient abnormal traffic labeled samples, transfer learning approach is used. In the proposed data-driven-based method the deep learning model is trained by traffic frequency features, which makes our model robust against AGC's parameters uncertainties and modeling nonlinearities.

## I. INTRODUCTION

AUTOMATIC generation control (AGC) system is an appealing target of cyber-physical attacks that cause either service interruptions or infrastructural damages. The vulnerabilities of AGC systems have been described in the literature [1, 2, 3, 4]. Anomaly detection in AGC systems is achieved via different approaches in the literature that can be divided into model-based and data-driven-based [5].

Many model-based algorithms were presented for anomaly detection in AGC systems. As an instance, in [6] kernel density estimation was used to detect false data injection attacks (FDIA). The authors in [7] utilized the Lyapunov stability theory to overcome FDIA in AGC systems. Threshold testing and a load forecasting-based algorithm are used in detecting FDIA in [8] and [9], respectively. Although these model-based methods offer competent detection methodologies, the need for accurate system models and parameters restricts the applicability of these algorithms in the actual networks [4].

To overcome the aforementioned issues, data-driven detection algorithms are used and developed. For example, the authors of [10] developed a detection algorithm based on a multilayer perception classifier. In [11] the authors proposed a neural network-based Luenberger observer for anomaly detection. In [12] regression-based FDIA signal prediction is developed using long short-term memory networks. Although these algorithms are model-free and there is no need for precise system models, still there are many challenges to be solved such as the high dimensionality of measurements that may be challenging to fit. Besides, the limited attack scenarios used in the training process question the ability of these detection algorithms in identify new attacks that were not trained for [5].

In this paper, to overcome these problems, a frequency domain feature extraction from IF is proposed using wavelet packet analysis theory [13]. Then, the extracted traffic frequency features are fed to the ResNeSt structure for abnormal IF identification. Furthermore, to solve the problem of the lack of abnormal traffic labeling samples in the AGC system, a transfer learning approach is used to train the abnormal IF detectors using a wider range of internet data samples.

## II. ANOMALY IN AGC: A BACKGROUND

The AGC system is used to maintain the scheduled power exchange at the tie-lines basically for mitigating power fluctuations; and maintain the frequency of the power system at its nominal value. In the typical AGC systems the frequency measurements and tie-lines power over different communication/network platforms such as IEC-61850 standard, within the electrical substation, and IEC-60870-5 standard, between the substations. To address the security issues in the aforementioned protocols IEC-62351 standard is used. However, still several vulnerabilities (e.g. device access, proximity access and network access) exist with the widely spread architecture of the power grids. These include micro-probing and circuit bonding, meter GPS spoofing, DNS spoofing and fake access points.

## III. PROPOSED METHOD

Since in cyber systems anomalous traffic is masked by high-volume normal data flows, it is difficult to accurately detect abnormal data flow by analyzing network communication traffic only in the time domain. To solve this problem, a wavelet-based traffic frequency analysis is used in this paper. In our proposed method thirty-nine features are used in anomaly traffic detection. Attributes 1 to 13 are inherent attributes of the IFs, such as source address, destination address, protocol type, etc, while the attributes 14 and 24, are the frequency-domain features extracted by the wavelet packet decomposition. Features 25 to 39, are the packet statistics attributes, such as the number of connections to the same host in the past two time slots.

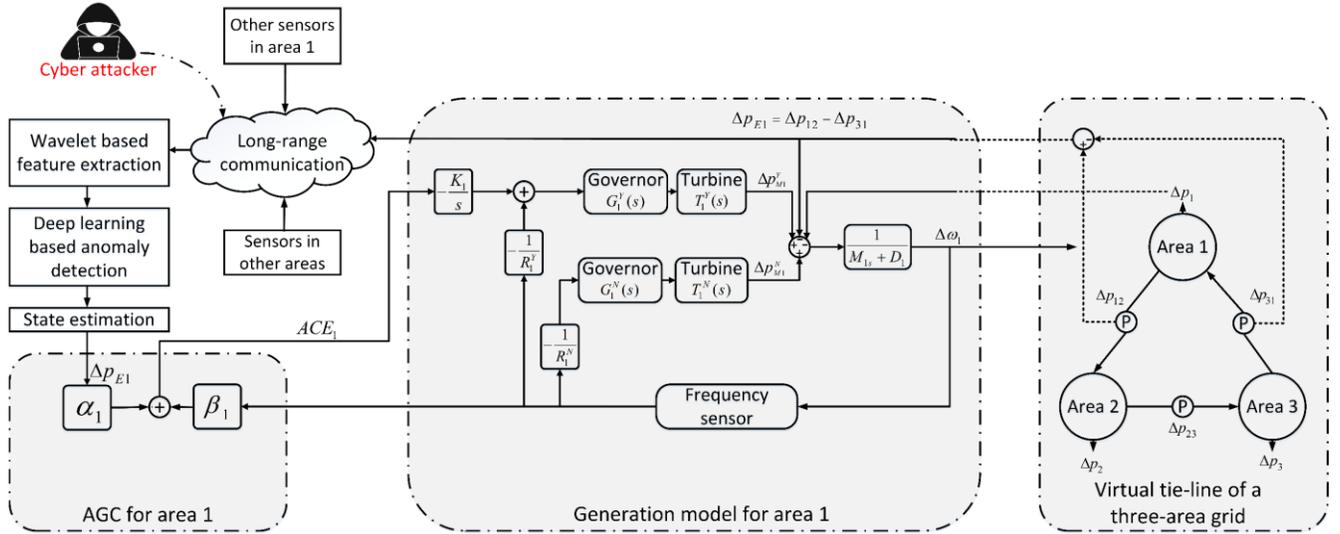

Fig. 1. Block diagram of AGC model

These thirty features cannot directly be fed into the ResNeSt model and should be pre-processed. In this regard, at first, the numerical features are normalized, basically to eliminate the effect of dimension differences, such as the number of variances of the upstream traffic, the number of variances of the downstream traffic, duration, packet size, etc. Then, the One-Hot encoding method is employed to map the symbolic features. Eventually, the thirty-nine features are encoded as 130-dimensional vectors and are converted into grayscale images, basically for using the powerful image processing capabilities of ResNeSt structure.

Eventually, to overcome the problem of the lack of attack scenarios (attack labels) in AGC systems, a transfer learning approach is used. Since the internet data has many public data sets of network anomalies and network attacks, they are used as the source domain, and the AGC system traffic is used as the target domain. As AGC system traffic and our used internet traffic share the same characteristics, the problem belongs to homogeneous transfer learning.

## IV. RESULTS

To evaluate the performance of the anomaly detection algorithms, the overall Accuracy (ACC), False Positive Rate (FPR), and Detection Rate (DR), are used. Table I shows the detection results of three different algorithms.

The results show that among the three deep learning approaches, ResNeSt proves to achieve the highest accuracy. This high accuracy can be explained by the combination of wavelet transform technique and residual neural network.

TABLE. 1. PERFORMANCE OF ANOMALY DETECTION ALGORITHMS

|         | ACC% | FPR% | DR%  |
|---------|------|------|------|
| DNN     | 94.7 | 7.1  | 93.0 |
| CNN     | 93.3 | 8.2  | 93.1 |
| RESNEST | 97.6 | 3.0  | 98.2 |